\documentclass{bmvc2k}

\usepackage{epsfig}
\usepackage{graphicx}
\usepackage{amsmath}
\usepackage{amssymb}
\usepackage{enumitem}
\usepackage{booktabs}

\DeclareMathOperator*{\argmax}{arg\,max}
\def\x{\boldsymbol{x}}

\title{Budget-Aware Activity Detection with A Recurrent Policy Network}

\addauthor{Behrooz Mahasseni}{behrooz.mahasseni@gmail.com}{1}
\addauthor{Xiaodong Yang}{xiaodongy@nvidia.com}{2}
\addauthor{Pavlo Molchanov}{pmolchanov@nvidia.com}{2}
\addauthor{Jan Kautz}{jkautz@nvidia.com}{2}

\addinstitution{Apple Inc.}
\addinstitution{NVIDIA Research}

\runninghead{Mahasseni, Yang, Molchanov, Kautz}{Budget-Aware Activity Detection}

\begin{document}

\maketitle

\begin{abstract}
In this paper, we address the challenging problem of efficient temporal activity detection in untrimmed long videos. While most recent work has focused and advanced the detection accuracy, the inference time can take seconds to minutes in processing each single video, which is too slow to be useful in real-world settings. This motivates the proposed budget-aware framework, which learns to perform activity detection by intelligently selecting a small subset of frames according to a specified time budget. We formulate this problem as a Markov decision process, and adopt a recurrent network to model the frame selection policy. We derive a recurrent policy gradient based approach to approximate the gradient of the non-decomposable and non-differentiable objective defined in our problem. In the extensive experiments, we achieve competitive detection accuracy, and more importantly, our approach is able to substantially reduce computation time and detect multiple activities with only 0.35s for each untrimmed long video.
\end{abstract}

\section{Introduction}
\label{sec:introduction}

Efficient temporal activity detection in untrimmed long videos is fundamental for intelligent video analytics including automatic categorizing, searching, indexing, segmentation, and retrieval of videos. This is a challenging problem as algorithms must (1) determine whether a specific activity occurs in an untrimmed video; (2) identify the temporal extent of each activity; and (3) maximize detection accuracy within a given time budget. In temporal activity detection, the most time consuming step is the execution of CNNs or hand-crafted feature extractors to every sliding window or proposal segment \cite{daps, fast-proposal, score-pyramid}, typically taking, e.g., seconds to minutes to process one video in THUMOS14 \cite{THUMOS14}. Unfortunately, this rules out the practical use of these methods for applications that require real-time and large-scale video processing. Although hardware solutions in some scenarios can help meet the constraints, it is equally important to establish a better understanding of how to achieve a maximal detection accuracy given the constraints on time and resource.

Recently, there is a fast growing interest in the research of temporal activity detection. Most existing work \cite{upc, oneata2014lear, singh2016untrimmed, wang2014action, score-pyramid} hinges on a large set of features and classifiers that exhaustively run over every time step at multiple temporal scales. This sliding window scheme is obviously computationally prohibitive for applications such as the ones running on mobile and embedded devices. To avoid such exhaustive evaluations, a number of action proposal algorithms \cite{daps, fast-proposal, shou2016temporal} have been proposed to produce a set of candidate temporal segments that are likely to contain a certain action. A separate classifier is then applied on these proposal segments for action classification. However, we argue that it is suboptimal to divide temporal activity detection into the two disjointed steps: proposal and classification. Moreover, the large number of proposal segments (e.g., thousands of proposals per video) is still unsatisfying in term of computational efficiency.   

\begin{figure*}
	\centering
	\includegraphics[width=0.85\linewidth]{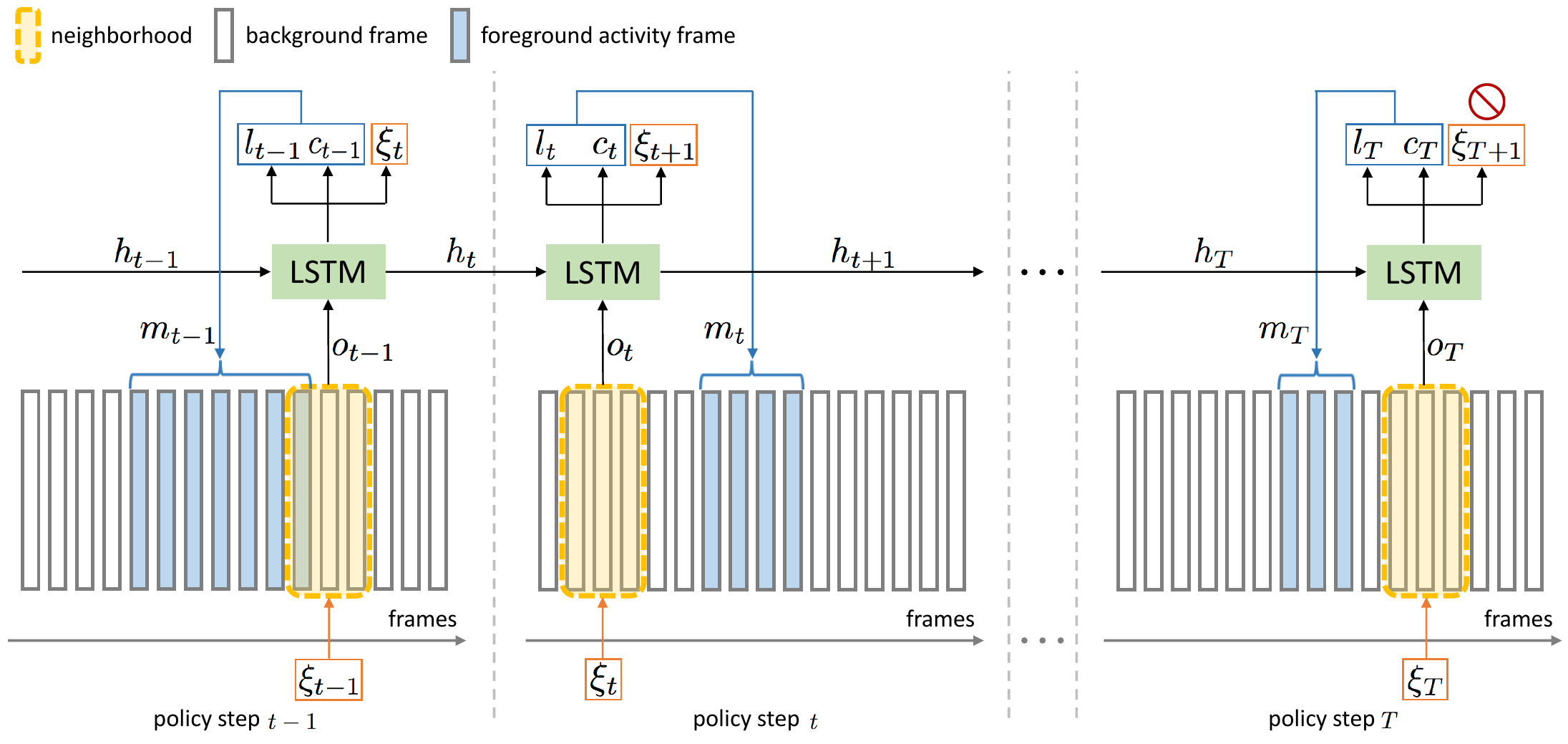}
	\caption {Given an untrimmed long video, at each step $t$ the policy has access to the local observation $o_t$ of a neighborhood centered around the current selected frame at $\xi_t$. For each step, the policy predicts a segment $m_t$ and produces three outputs: the temporal location $l_t$ (i.e., start and end) of the segment, the estimated class $c_t$ of the segment, and the next frame to observe at $\xi_{t + 1}$. The policy runs for $T$ steps according to a specified time budget.}
	\label{fig:block-diagram}
\end{figure*}

In this paper, we address this problem by introducing a fully end-to-end and budget-aware framework, which learns to optimally select a small number of video frames according to a time budget to perform temporal activity detection. We formalize our frame selection as a Markov decision process (MDP), and adopt a recurrent network to model the policy for selecting frames. We develop a policy gradient approach based on reinforcement learning to approximate the gradient of our non-decomposable and non-differentiable objective function. Figure \ref{fig:block-diagram} illustrates the detection process of our approach.

Our main contributions of this paper are as follows. First,
we achieve competitive accuracy by only 0.35s for each untrimmed minutes-long video, speeding up detection by orders of magnitude compared to the existing methods. Second, we present a fully end-to-end policy based model with a single training phase to handle the activity classes in their entirety. Third, to our knowledge, we provide the first approach to directly optimize the mAP criteria (i.e., the final evaluation metric) in the objective function for activity detection. 

\section{Related Work}
\label{sec:related-work}

A large family of the research in video activity understanding is about activity classification, which provides useful tools for temporal activity detection, such as the two-stream networks with two separate CNNs operating on the color and optical flow modalities \cite{two-stream}, and the C3D using 3D-CNN to process short video clips \cite{c3d}. RNNs can be also applied with CNNs to model the temporal dynamics and handle variable-length video sequences \cite{prernn}.    

Another active research line is the spatio-temporal action detection, which focuses on localizing action regions over consecutive video frames. A number of methods have been proposed, from the spatio-temporal proposals such as supervoxels \cite{mshah}, the frame-level object detection followed by a linking or tracking algorithm \cite{action-tube}, to the recent video SSD approach \cite{video-ssd}. However, these methods are mostly applied on short video snippets, in contrast, temporal activity detection targets at untrimmed videos that involve complex actions, objects and scenes evolving over long sequences. Therefore, efficient inference under a certain time budget is much in demand for the temporal activity detection task. 

A majority of the existing approaches \cite{upc, oneata2014lear, singh2016untrimmed, wang2014action, score-pyramid} for temporal activity detection focus on extracting various features to represent sliding windows and subsequently classifying them with SVMs trained on the multiple features. Alternative proposal based methods can be used to generate action segments to replace the exhaustive sliding windows. A sparse learning based framework is presented in \cite{fast-proposal} to produce segment proposals with high recall. Escorcia et al. \cite{daps} introduce a proposal algorithm based on C3D and LSTM. Shou et al. \cite{shou2016temporal} propose a 3D-CNN based multi-stage framework with three different networks for proposal, classification and localization. A convolutional-de-convolutional network is further introduced in \cite{DBLP:journals/corr/ShouCZMC17} to boost the precision for temporal boundaries of proposal segments. These proposal based detection methods are by nature stage-wised, and therefore are not end-to-end trainable. R-C3D is recently developed in \cite{r-c3d} to save computation through sharing convolutional features between proposal and classification pipelines. 

Yeung et al. \cite{yeung2015end} present an end-to-end learning method to predict temporal boundaries from raw videos. In that work, a recurrent attention model is learned to select a subset of frames to interact with, and maintains high detection accuracy. Our work differs from \cite{yeung2015end} mainly in: (1) unlike their binary model specifically trained for each action class, our approach is able to handle multiple classes; (2) rather than using a separate emission signal to identify foreground segments, we consider all predicted outputs as valid segments since we include the background as an additional class; (3) their reward function is designed to maximize true positives and minimize false positives, while our retrieval loss is directly defined on the mAP; and (4) instead of using two schemes (i.e., back-propagation for candidate detection and REINFORCE for prediction indicator and next observation) to train their learning agent, we employ a unified recurrent policy gradient to train the entire policy altogether.

\section{Problem Formulation}
\label{sec:problem-formulation}

Given a video $v$ and a set of activity labels $\mathcal{L}$ , our goal is to predict for each frame a single label from $\mathcal{L}$. We call each temporal extent consisting of consecutive frames with the same label a semantic temporal segment. As stated in Section~\ref{sec:introduction}, given a limited time budget, it is infeasible to process every single frame in a video. So we aim to detect and classify the foreground segments by only observing a small subset of video frames $\x \in v$. 

Assuming limited access to the frames of $v$, finding the optimal frame subset $\x$ is inherently a sequential decision making task. Accordingly, we draw on ideas from reinforcement learning---an area that focuses on learning for sequential decision making problems. Our aim is to learn a policy $\pi$, parameterized by $\theta$, to sequentially select the frames from $v$ and form the subset $\x$. Alongside the selection process, $\pi$ outputs the current belief about the foreground segment and the associated class label. This sequential decision making process intuitively resembles how humans search activities in a video, i.e., iteratively refine our estimated temporal boundaries by sequentially choosing a few frames to observe.         

Let $\mathbb{G}$ denote the ground truth segments in $v$, and $\mathbb{M}^x$ be the set of estimated semantic temporal segments from observing $\x$. We define the deterministic indicator $\mathbb{I}_{m, g}$ to identify whether an estimated segment $m \in \mathbb{M}^x$ is assigned to a ground truth segment $g \in \mathbb{G}$:
\begin{eqnarray}
\displaystyle \mathbb{I}_{m, g} = \begin{cases} 
1 & g =\argmax_{g^\prime \in \mathbb{G}} \alpha (m ,g^\prime) \hspace{3pt} \text{subject to } \alpha > 0, \\
0 & \text{otherwise},
\end{cases}
\end{eqnarray}
\noindent where $\alpha$ is the intersection over union (IoU). Let $c_m$ and $c_g$ indicate the probability distribution and one-hot representation of the associated class labels for segments $m$ and $g$. For a subset of selected frames $\x$ and a set of predicted segments $\mathbb{M}^x$, our loss is defined as:
\begin{eqnarray}\label{eq:loss}
L_\theta = \sum_{m \in \mathbb{M}^x}\sum_{g \in \mathbb{G}} \mathbb{I}_{m, g} \Big[ \lambda_{c} \Delta_{cls} (c_m, c_g) + \lambda_{l} \Delta_{loc} (l_m, l_{g}) \Big] + \lambda_{r} \Delta_{ret}(\mathbb{M}^x, \mathbb{G}), 
\end{eqnarray}
\noindent where $\Delta_{cls}$ is the multi-class classification error, $\Delta_{loc}$ is the localization error with $l_m$ and $l_g$ identifying the locations of segments $m$ and $g$, and $\Delta_{ret}$ is the segment retrieval error. The most important property of $\Delta_{ret}$ is that while it encourages the model to detect all foreground segments, it also discourages the model from producing many false positives. 

We now explain how to formulate each individual error defined in Eq.~(\ref{eq:loss}). In contrast to using a binary classification loss as in \cite{yeung2015end}, we employ a multi-class cross-entropy loss $\Delta_{cls} = - c_g \log {c_m}$. Unlike \cite{yeung2015end} which penalizes the localization based on the absolute error, we believe this loss should also depend on the duration of a segment, i.e., the same amount of absolute error should be treated differently for short and long intervals. Intuitively, this means that if the policy makes a small error for a short segment this error should be considered relatively large, otherwise the algorithm would ignore the small segments. With this intention, we define $\Delta_{loc} (l_m, l_g) = \zeta(g) \times \|(m_s, m_e), (g_s , g_e) \|$, where $\zeta(g)$ is a scaling factor which depends on the length of segment $g$, $\|\cdot\|$ is the distance between two segments, $m_s$ and $m_e$ are the start and end of segment $m$, similar for segment $g$. To define the segment retrieval loss $\Delta_{ret}(\mathbb{M}^x, \mathbb{G})$, we use the mAP criteria, where mean is over different class labels, and AP for each individual class is defined as $\text{AP}(\mathbb{M}^x, \mathbb{G}) =  \sum_i \text{Prec}(\mathbb{M}^x(i), \mathbb{G}) \times \Delta_\text{Recall}$, where $\mathbb{M}^x(i)$ is the subset of $\mathbb{M}^x$ till the $i$th segment ranked by the overlap with ground truth, $\text{Prec}(\cdot)$ is the precision of detection, and $\Delta_\text{Recall}$ is the change of recall from previous subset. Given a training set of $N$ videos $\left \{ v_1, \cdots, v_N \right \}$, our goal is to find $\theta$ that minimizes: 
\begin{equation}
\theta^* =\arg\min_{\theta} \big [\mathbb{E}(L_\theta) \approx \frac{1}{N}\sum_{n=1}^N L_\theta(\mathbb{M}^x_n, \mathbb{G}_n) \big].
\label{eq:expected_loss}
\end{equation}
\noindent Unfortunately, the standard back-propagation is not applicable to learn the parameters in Eq.~(\ref{eq:expected_loss}), as the objective function in Eq.~(\ref{eq:loss}) contains the non-differentiable components. This is mainly due to the non-decomposable AP, as well as the sequential decision making process in selecting video frames. In order to solve this difficulty, we reformulate our problem as a reinforcement learning problem, which allows us to define an equivalent reward function to the original objective function.


\section{Recurrent Policy Network}\label{sec:problem-statement}
\subsection{Policy Representation}
\label{sec:policy-representation}

Our activity detection agent makes a sequence of predictions based on the local information from the most recent observed frame. At each step, the policy produces three outputs including the estimate of start and end of the current potential temporal segment, the prediction of class label associated with the segment, and the next frame to observe. Unlike the binary model used in \cite{yeung2015end}, our approach enables us to define a multi-class classifier, which means that we only need to train a single policy rather than training multiple different policies. Note that this also allows us to avoid the binary prediction indicator signal used in \cite{yeung2015end}, since we can directly discard those segments predicted with the background label. 

Due to the local observation at each step, the policy has no access to the global state (i.e., the entire video). This resembles the partially observable Markov decision process (POMDP), which assumes that despite the existence of a global state, for practical reasons an agent does not have a full observation of the global state. We adopt the recurrent policy gradient approach \cite{wierstra2010recurrent} to maintain an approximate belief of the current state $s_t$ by LSTM.

Particularly, suppose at step $t$ the current frame is $i$, the policy $\pi$ makes a decision based on (1) the local information of a neighborhood $\mathcal{N}_i$ centered around $i$ and (2) the history of previous observations. We capture the local information through an observation feature  $o_t = [\psi(\mathcal{N}_i), \phi(\mathcal{N}_i), \xi_t]$, where $\psi(\mathcal{N}_i)$ is an indicator vector that identifies whether each frame in $\mathcal{N}_i$ has been previously selected, $\phi({\mathcal{N}_j})$ is the average of per-class confidence predicted in $\mathcal{N}_i$ , and $\xi_t \in [0,1]$ is the normalized location of the current frame at step $t$. The inclusion of $\xi_t$ is helpful in encouraging the policy to cover broader video content. During our experiments, we find that excluding $\xi_t$ results in a considerable number of over-selection of frames. Note that for $\phi$, we compute the averaged confidence of estimated segments, which share the frames in $\mathcal{N}_i$. As for the history of the decision makings, we use the hidden state $h_{t-1}$ of LSTM to maintain the context of previous observations up to step $t$.

To summarize, the global state at step $t$  is approximated by the internal state $h_t$ of LSTM, which depends on the current observation $o_t$ and the previous state $h_{t-1}$. Given $h_t$ the outputs of the policy $\pi$ are $\nu_t = \left[ l_t, c_t, \xi_{t+1} \right]$: (1) the location $l_t$ of an estimated temporal segment, (2) the probability distribution over activity class labels $c_t$, and (3) the location of the next observation $\xi_{t+1}$. Note that our formulation allows the policy to perform both forward and backward selections. In order to further improve the exploration at training phase, instead of directly using $\xi_{t+1}$, the next selected location is sampled from a Gausssian distribution with a mean equal to $\xi_{t+1}$ and a fixed variance (e.g., 0.18 used in our experiments). 

\subsection{Policy Learning}
\label{sec:policy-learning}

Our goal of policy learning is to jointly optimize the parameters of $\pi$ by minimizing the loss of a sequence of policy actions as defined in Eq.~\eqref{eq:loss}. These actions are taken from the initial state $s_0$, when no frames are selected, until the final state $s_T$, where $T$ is the number of steps specified according to a time budget. 

The main difficulty in policy learning is that the estimated temporal segments $\mathbb{M}^x$ for a video are computed through a sequence of policy decisions, resulting in a non-decomposable and non-differentiable objective function. Moreover, a decision that the policy makes at any step depends on the history of decisions that the policy has made in previous steps, and also impacts the decisions available to the policy in the future. 
Among the potential algorithms for addressing similar POMDP problems \cite{cho2017robust,egorov2017pomdps,wierstra2010recurrent,williams1992simple}, we adopt the recurrent policy gradient approach \cite{wierstra2010recurrent}, which provides better theoretical bounds on learning objective to approximate the gradients of our non-decomposable and non-differentiable objective function, so that the policy can be efficiently learned with stochastic gradient descent.

To follow the general reinforcement learning formulation, let $r$ be the immediate reward associated with a state $s_t$. Since $s_t \approx h_t$ in our policy, we define $r$ as $r(h_t) =  L_\theta(\mathbb{M}^x_{t-1}, \mathbb{G}) - L_\theta(\mathbb{M}^x_t, \mathbb{G})$, where $L_\theta$ is the loss associated with a set of estimated temporal segments as defined in Eq.~(\ref{eq:loss}). Intuitively, $r(h_t)$ states that the policy earns an immediate reward equal to the decrease in the temporal segmentation error achieved by selecting an observed frame, or pays a penalty if the temporal segmentation error increases. Let $R(H_t)$ be the discounted accumulated reward starting from the state $s_t$ and continuing the policy up to the final state $s_T$: $R(H_t) = \sum_{t^\prime = t}^T \tau^ {t^\prime - t} r(h_{t^\prime})$, where $H_t = \{h_t, ..., h_T\}$ represents the history of hidden states in LSTM, and $\tau\in(0,1)$ is the discount factor. $H_0$ can be interpreted as the trajectory of observations for a sample run of the policy from the initial state. For notational simplicity, we use $H$ for $H_0$ in the rest of this paper. The goal of policy learning is transformed to find the parameters $\theta^*$ to maximize $J(\theta)$ which is defined as:
\begin{equation}
\label{eq:acc_reward}
J(\theta) = \mathbb{E}[R(H)] = \int p(H|\theta) R_\theta(H) dH,
\end{equation}
\noindent where $p(H |\theta)$ is the probability of observing a sequence of hidden states $H$, given a policy $\pi$ defined by the parameters $\theta$. It can be shown that maximizing $J(\theta)$ implicitly minimizes $L_\theta$ along the trajectory of policy executions. We now derive how to compute the gradient with respect to the policy parameters $\nabla_{\theta}J$, which is given by: 
\begin{equation}
\label{eq:gradient1}
\nabla_{\theta}J = \int \big[\nabla_{\theta} p(H|\theta) R_\theta(H) + p(H|\theta) \nabla_{\theta} R_\theta(H)\big] dH.
\end{equation}
\noindent Note that given the sequence of hidden states $H$, which determines the history of selected frames, the reward function does not depend on the policy parameters, yielding $\nabla_{\theta} R_{\theta}(H) = 0$. To further simplify Eq.~\eqref{eq:gradient1}, we need to define $\nabla_{\theta} p(H|\theta)$. So we first factorize $p(H|\theta)$ as $p(H|\theta) = p(h_0) \prod_{t=1}^T p(h_t | h_{t-1}) \pi(\nu_t | h_{t-1}, o_t)$, where the same notation $\pi$ is used to denote the output of the policy. Based on this we have: $\log p(H|\theta) = \text{const} + \sum_{t=1}^T \log \pi(\nu_t | h_{t-1}, o_t)$, where the first term is a sum over the log of $p(h_t| h_{t-1})$, a constant with respect to $\theta$. This therefore results in the following gradient: $\nabla_{\theta} \log  p(H|\theta) = \sum_{t=1}^T \nabla_\theta \log \pi(\nu_t | h_{t-1}, o_t)$.

It is common to use the Monte-Carlo integration to approximate the integration over the probability of observing a sequence of hidden states. Specifically, the approximate gradient is computed by running the current policy on $N$ training videos to generate $N$ trajectories. Combining aforementioned derivations and Eq.~(\ref{eq:gradient1}), we can obtain the approximate gradient:
\begin{eqnarray}
\label{eq:gradient1-2}
\nabla_\theta J \approx  \frac{1}{N} \sum_{n=1}^N \sum_{t=1}^T \big[\nabla_{\theta} \log \pi(\nu_t^n | h_{t-1}^n, o_t^n) R_{\theta}(h_t^n) \big].
\end{eqnarray}
\noindent Since the policy gradient methods usually suffer from the high variance of gradient estimates, we follow the common practice used in \cite{williams1992simple} to subtract a bias from the expected reward $R$. However, rather than taking a constant bias, we set the bias value to be the reward obtained from a random selection policy.


\section{Experiments}


In this section, we extensively evaluate our approach on the two benchmark datasets: THUMOS14 \cite{THUMOS14} and ActivityNet \cite{caba2015activitynet}. Experimental results demonstrate that our approach is able to substantially reduce computational time, and meanwhile provides competitive detection accuracy under varying time budgets. In the appendix, we explain how to calculate the detection speeds of different methods, and present the illustration of the learned policy for frame selection, as well as the examples of our model's predictions. 

We use the pre-trained VGG16 \cite{simonyan2014very} on ImageNet as our backbone CNN, and fine-tune the network on each dataset. We take the layer \texttt{fc7} of VGG16 as the per-frame feature. Our policy is based on a two-layer LSTM, and each layer has 1024 hidden units. If not otherwise specified, our policy takes $T = 6$ steps, which we believe are efficient enough to meet a reasonable time budget constraint. We empirically set the weights in our loss function of Eq.~\eqref{eq:loss} as $\lambda_{c} = \lambda_{l} = 1.0$ and $\lambda_{r} = 0.5$. We set the batch size to 128, and train each model for 100 epochs using Adam with the default parameters. We implement our networks in TensorFlow and perform experiments on a single NVIDIA Tesla P100 GPU.

\begin{figure}[t]
	\centering
	\includegraphics[width=\linewidth]{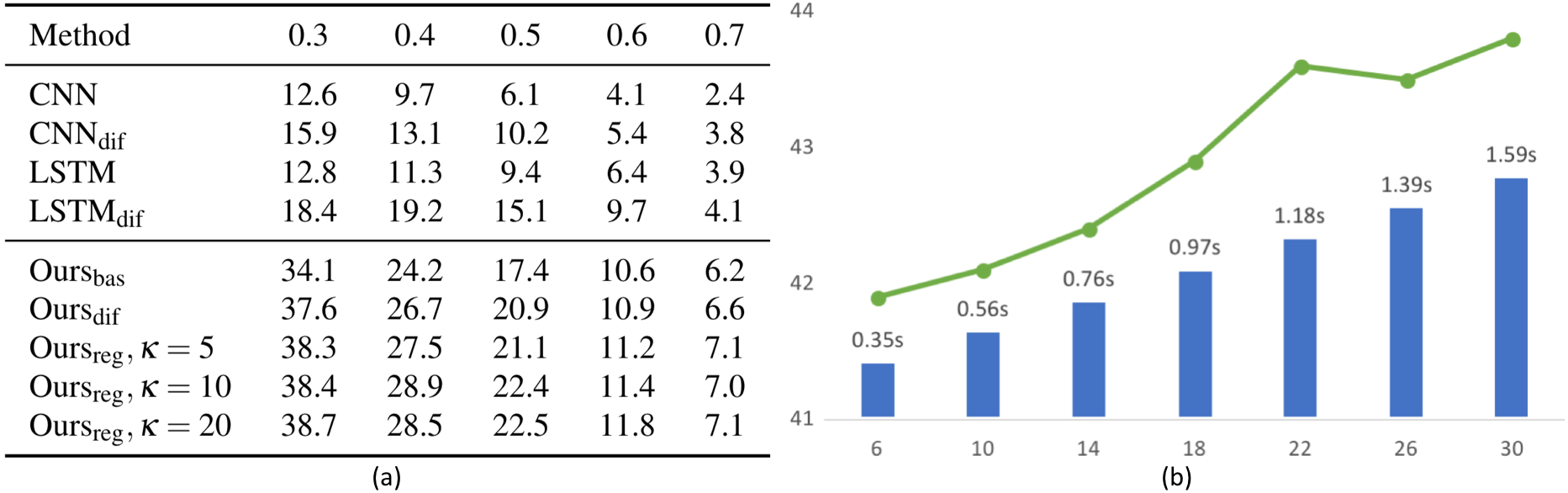}
	\caption {(a) Comparison of the detection accuracy of baselines and variations of our models under different IoU thresholds $\alpha$ on THUMOS14. (b) Comparison of the detection accuracy (green dots) and the speed (blue bars) with different policy steps on ActivityNet.}
	\label{fig:table_figure}
\end{figure}


\subsection{Baselines and Variations of Our Model}


To provide a better insight, we first study the different configurations of baseline methods and our model. We define the following two important baselines, each of them outputs a class label for a single frame including the background class, followed by a non-maximum suppression (NMS) post-processing to aggregate the class probabilities of single frames. 

\begin{itemize}[noitemsep,topsep=0pt]
	\item \textbf{CNN with NMS}: VGG16 is fine-tuned on the single frames of each dataset and used to perform per-frame classification.
	\item \textbf{LSTM with NMS}: According to the setting in our model, we train LSTM on top of the layer \texttt{fc7} of VGG16, and a prediction is made for each single frame. 
\end{itemize}

\noindent We can improve our base model mainly in the two ways and define the multiple variations of our model as follows. 

\begin{itemize}[noitemsep,topsep=0pt]
	\item $\textbf{Ours}_{\textbf{bas}}$ is the base model defined in Sec.~\ref{sec:policy-representation}.
	\item $\textbf{Ours}_{\textbf{dif}}$ provides the simple pixel-level frame difference between consecutive frames to roughly capture motion cues. Although we can incorporate optical flow, given the fact that optical flow is computationally expensive, and the motivation of this work is for fast inference, we believe that the pixel-level frame subtraction is a reasonable compromise. We use early-fusion to concatenate RGB channels of the original frame and the frame difference as a composite input to VGG16. 
	\item $\textbf{Ours}_{\textbf{reg}}$ performs a simple post-processing to refine the policy output. A simple linear regressor is trained to refine the boundaries of detected temporal segments by using the current temporal extend of a segment, and the $\kappa$ uniformly sampled frames along with their pixel-level frame differences within this segment. 
\end{itemize}


\subsection{Results on THUMOS14}


We follow the standard experimental setting on THUMOS14~\cite{THUMOS14}. We first present the ablation study to understand the contribution of each individual component in our model. As expected and shown in Figure~\ref{fig:table_figure}(a), providing the additional simple frame difference to capture coarse motion cues improves the accuracy of all baselines and our models. The baseline recurrent models with LSTM are found to produce better results than CNN. 
All of our ablation models significantly outperform the baselines, quantifying the contributions of our proposed approach. In particular, our results are generated by only observing 6 policy-selected frames, far more efficient than the baselines that have to densely go through all video frames.      

It is interesting to observe that the simple linear regression based post-processing with $\kappa$ uniformly sampled frames from estimated segments helps in refining the temporal boundaries in Figure~\ref{fig:table_figure}(a). We conjecture that this is due to the fact that our policy is allowed to observe frames in a temporally inconsistent way (i.e., selecting frames in a mixed forward and backward fashion). LSTM thus tends to smooth out the features to some extent during this process. We hypothesize that observing the $\kappa$ sampled frames in the simple regression provides a temporally consistent description complementary to the averaged latent representation that lacks temporal consistency in the policy. We also evaluate the impact of the number of sampled frames $\kappa$ to the regression. As shown in Figure~\ref{fig:table_figure}(a), we only observe marginal gains when sampling over 10 frames, which also implies that our policy has already learned to select fairly representative frames to perform the detection.   

\begin{table*}[h]
	\small
	\centering
	\begin{tabular}{lrccccc}
		\toprule
		Method & Time (s) & $\alpha = 0.3$ & $\alpha = 0.4$ & $\alpha = 0.5$ & $\alpha = 0.6$ & $\alpha = 0.7$ \\
		\midrule
        LEAR14 \cite{oneata2014lear} & $>$ 108 & 28.8& 21.8& 15.0& 8.5& 3.2\\
        CUHK14 \cite{wang2014action}& $>$ 108 & 14.6& 12.1& 8.5& 4.7& 1.5\\
        Pyramid of Scores \cite{score-pyramid} & $>$ 108 & 33.6 & 26.1 & 18.8 & - & - \\
        Fast Temporal Proposals \cite{fast-proposal} & 108 & - & - & 13.5 & - & - \\
        S-CNN \cite{shou2016temporal}& 92 & 36.3 &28.7 & 19.0 &10.3& 5.3\\
        CDC \cite{DBLP:journals/corr/ShouCZMC17} & 92 & 40.1 & 29.4  & 23.3 & 13.1 & 7.9 \\
        DAPs \cite{daps} & 41 & - & - & 13.9 & - & - \\
        Language Model \cite{language-model} & 17 & 30.0 & 23.2 & 15.2 & - & - \\
        R-C3D \cite{r-c3d} & 5.3 & 44.8 & 35.6 & 28.9 & - & - \\
		Glimpses \cite{yeung2015end} & 4.9 & 36.0  & 26.4  & 17.1  & - & - \\
        \midrule
		Ours & \textbf{0.35} & 38.4 & 28.9 & 22.4 & 11.4 & 7.0 \\
		\bottomrule
	\end{tabular}
    \vspace{10pt}
	\caption{Comparison of our approach and the state-of-the-art methods in the approximate computation time (s) to process each video and the detection accuracy (mAP) over different IoU thresholds on THUMOS14.}
	\label{tab:thumos-soa}
\end{table*}

\begin{figure}[t]
	\centering
	\includegraphics[width=\linewidth]{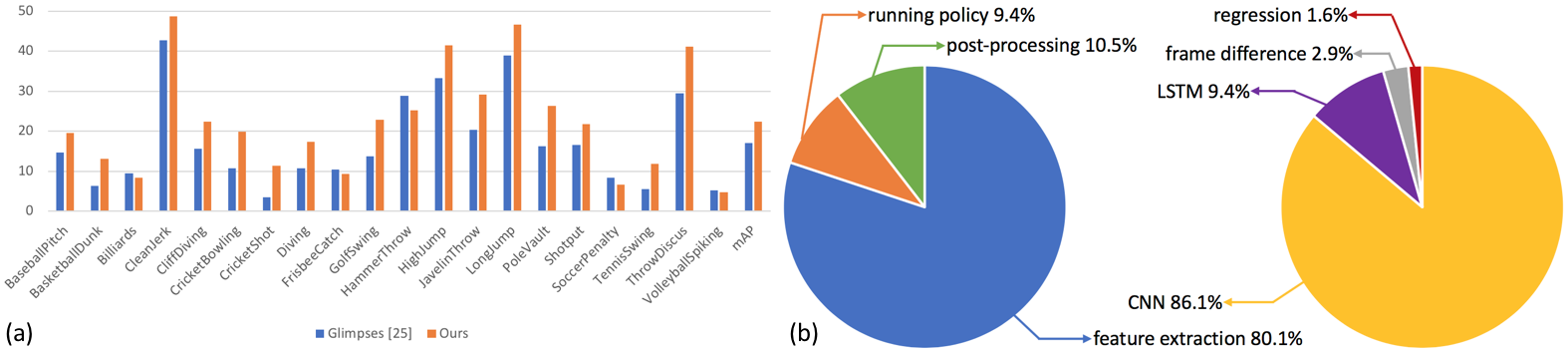}
	\caption {(a) Comparison of the per-class breakdown AP at $\alpha = 0.5$ on THUMOS14. (b) Analysis of computational time of our approach: percentage of time spent on each major algorithm step (left) and computation component (right) to perform detection.}
	\label{fig:time}
\end{figure}

We then compare with the state-of-the-art methods in Table~\ref{tab:thumos-soa}. Our approach achieves competitive detection accuracy under various IoU thresholds, and more importantly, we perform detection in only 0.35s for each untrimmed long video. This is orders of magnitude faster than most other competing algorithms relying on sliding windows or segment proposals. While R-C3D produces superior accuracy on this dataset, we significantly outperform R-C3D on ActivityNet (see Table~\ref{tab:activitynet_allt}), indicating the advantage of our approach to handle more complex activities. We specifically compare the per-class breakdown AP of our model and the glimpses method \cite{yeung2015end} that also exhibits efficient inference for each binary detection. As shown in Figure~\ref{fig:time}(a), our approach largely outperforms \cite{yeung2015end} in 15 out of 20 classes, and by 5.3\% in overall mAP. Notably, our method is a unified model to handle all classes, while \cite{yeung2015end} is a binary model that needs to be trained for each specific class of the 20 actions. 

We also provide the in-depth analysis of the computational costs of our approach. Figure~\ref{fig:time}(b) shows the percentage of time for each major algorithm step and computation component. Our policy is quite efficient to run, takes only $9.4\%$ of the time. The feature extraction that involves applying CNN to multiple frames dominates the computations, consuming $80.1\%$ of the time, which again highlights the importance of effective frame selection to reduce the computational burden. In addition, it uses only $2.9\%$ of the time to compute the frame differences, which can provide coarse but useful motion cues. 

\begin{table*}[t]
	\small
	\centering
	\begin{tabular}{lrcccc}
		\toprule
		Method & Time (s) & $\alpha = 0.5$ & $\alpha = 0.75$ & $\alpha = 0.95$ & ave-mAP \\
		\midrule
		UTS16 \cite{uts}& $>$ 930 & 45.1 & 4.1 & 0.0 & 16.4\\
        CDC \cite{DBLP:journals/corr/ShouCZMC17} & $>$ 930 & 45.3 & 26.0 & 0.2 & 23.8 \\
        UPC16 \cite{upc} & 11 & 22.5 & - & - & - \\
        OBU16 \cite{singh2016untrimmed}& 10 & 22.7 & 10.8 & 0.3 & 11.3 \\
        R-C3D \cite{r-c3d} & 3.2 & 26.8 & - & - & - \\
		\midrule
        Ours & \textbf{0.35} & 41.9 & 22.6 & 0.1 & 21.5 \\
		\bottomrule
	\end{tabular}
	\vspace{10pt}
	\caption{Comparison of our approach and the state-of-the-art methods in the approximate computation time (s) to process each video and the detection accuracy (mAP) over different IoU thresholds on ActivityNet.}
	\label{tab:activitynet_allt}
\end{table*}  

\subsection{Results on ActivityNet}

Compared to THUMOS14, ActivityNet~\cite{caba2015activitynet} contains high-level semantics with complex actions, objects and scenes in videos, and is much larger in number of activities and amount of videos. Based on this large-scale dataset, we first evaluate our model with different policy steps, according to different time budgets the detection system can afford. As shown in Figure~\ref{fig:table_figure}(b), when the policy steps move up from 6 to 30 by increasing the time budget from $0.35$s to $1.59$s, the detection accuracy is improved by about $2.0\%$ under $\alpha = 0.5$.   

Finally, we compare our approach with the state-of-the-art methods in Table~\ref{tab:activitynet_allt}. Similar to the results on THUMOS14, our approach substantially reduces the detection time by orders of magnitude compared to other methods. While CDC provides very competitive accuracy, it relies on the detection results of UTS16, i.e., CDC is primarily used to refine the predicted temporal boundaries of UTS16. If directly applying CDC on raw videos, the accuracy drops to around $15.0\%$ at $\alpha = 0.5$. UTS16 is sliding window based, and requires multiple expensive feature extractions including iDT, C3D, ResNet152, and InceptionV3. We achieve significant improvement over the most recent state-of-the-art method R-C3D, demonstrating the superiority of our approach to tackle the more complex activities. Since the glimpses method \cite{yeung2015end} is a binary model and their detection result on the entire 200 classes of this dataset is not provided, we train our policy on the same two subsets (i.e., sports and work) as \cite{yeung2015end} for fair comparisons. More details are provided in the appendix. 
 
\section{Conclusion}

We have presented a fully end-to-end approach for the challenging problem of efficient temporal activity detection. We formulate the budget-aware inference for this problem to optimally select a small subset of frames within the MDP framework. We propose the LSTM based policy model to handle the whole activity classes by a single training phase. A policy gradient is developed to approximate the gradient of our non-decomposable and non-differentiable objective. Experiments demonstrate that our approach brings substantial time saving and maintains competitive detection accuracy. This provides a practical solution for many applications that require tight runtime constraints and limited on-device computations.

\bibliography{egbib}

\section*{Appendix}

\setcounter{section}{0}
\renewcommand\thesection{\Alph{section}}

In this appendix, Section~\ref{sec:time} summarizes how to calculate the approximate detection time of different methods. Section~\ref{sec:policy} illustrates the frame selection and prediction process of our approach. Section~\ref{sec:examples} demonstrates the examples of detected activities by our approach. Section~\ref{sec:results} provides more results on ActivityNet. 

\section{Estimation of Detection Time}
\label{sec:time}

It is challenging to decide the precise computation time of the different methods due to (1) lack of computational information in the literature, and (2) multiple complex stages involved. We adopt the following strategies to estimate the approximate computation time for them: (1) directly use the speed or time if it is reported in the related papers; (2) use the speed or time of our components (e.g., CNN and LSTM) to infer others; (3) use the processing bottleneck that dominates the computational costs (e.g., extraction of hand-crafted features) to approximate the overall time.

\subsection{THUMOS14}

There are 20 action classes and 1574 videos in the testing set of THUMOS14. Each video has 5507 frames on average. 

\begin{itemize}
\item \textbf{Ours} is provided in the analysis of computational costs of Figure 3(b) in the paper. The detailed computation time of each component is: forward pass of VGG16 is 3.0ms for each frame on the GPU; execution time of 2-layer LSTM at each step is 5.4ms on the GPU; pixel-level frame difference is 0.1ms per frame on the CPU; linear regression is 5.5ms on the CPU. We use 6 steps to run our policy for each video, 15 frames as the neighborhood of each selected frame, and 10 uniformly sampled frames for regression. So the overall computation time for each video is: $(0.1\times15\times6) + (3.0\times15\times6) + (5.4\times6) + (0.1\times10) + (3.0\times10) + 5.5 \approx 348$ms $\approx 0.35$s. 

\item \textbf{Glimpses}~\cite{yeung2015end}  downsamples videos by 5 (i.e., to 5fps) for this dataset, and observes $2\%$ (or less) of video frames for inference. Since it is a binary model, we need to train and run 20 models to detect all classes of this dataset. Based on the time of our approach, we can infer the computation time of each component in [25]: forward time of VGG16 is 3.0ms for each frame on the GPU; execution time of 3-layer LSTM at each observation is 8.2ms on the GPU. So the average computation time to detect the whole classes for each videos is: $5507 \div 5 \times 0.02 \times (3.0 + 8.2) \times 20 \approx 4932$ms $\approx 4.9$s.

\item \textbf{R-C3D}~\cite{r-c3d} runs at 569fps on an NVIDIA Titan X Maxwell GPU and 1030fps on an NVIDIA Titan X Pascal GPU. So the average computation time to predict each video is: $5507 \div 569 \approx 9.7$s on Titan X Maxwell, and $5507 \div 1030 \approx 5.3$s on Titan X Pascal.

\item \textbf{Language Model}~\cite{language-model} takes 7.5h on the CPU with eight 1.2GHz cores for inference on this dataset. So the average computation time for each video is: $7.5 \times 60 \times 60 \div 1574 \approx 17$s. 

\item \textbf{DAPs}~\cite{daps} integrates C3D and LSTM, and runs at 134.1fps on the GPU. So the average computation time to predict each video is: $5507 \div 134.1 \approx 41$s.

\item \textbf{S-CNN}~\cite{shou2016temporal} employs the proposal and localization networks for inference, and runs at 60fps on the GPU. So the average computation time for each videos is: $5507 \div 60 \approx 92$s.  

\item \textbf{CDC}~\cite{DBLP:journals/corr/ShouCZMC17} alone runs at 500fps on the GPU to refine predicted temporal boundaries. However, on this dataset, it relies on the proposals generated by S-CNN~\cite{shou2016temporal}, which dominates the computational costs. So the overall computation time including segment proposal and prediction refinement for each video can be approximated by that of S-CNN: 92s.       

\item \textbf{Fast Temporal Proposal}~\cite{fast-proposal} downsamples videos by 5 and runs at 10.2fps on the CPU. So the average computation time for each video is: $5507 \div 5  \div 10.2 \approx 108$s. 

\item \textbf{Pyramid of Scores}~\cite{score-pyramid} employs a sliding window strategy, and requires time-consuming feature extraction, e.g., 21d to extract the iDT features on the CPU, 20h for feature encoding on the CPU, etc. So the average computation time to predict each video is much longer than the proposal based method in~\cite{fast-proposal}: $> 108$s.     

\item \textbf{CUHK14}~\cite{wang2014action} is also a sliding window based approach. It extracts deep features by AlexNet, and iDT features running at 3.5fps on the CPU. Additional computations include FV encoding of iDT features and SVM classification. So the average computation time to process each video is also much longer than the proposal based method in~\cite{fast-proposal}: $> 108$s.   

\item \textbf{LEAR14}~\cite{oneata2014lear} is also based on  sliding windows with computationally expensive feature extractions: SIFT, iDT, color features, FV encoding, deep features by AlexNet, as well as acoustic features such as MFCC and ASR. So the average computation time to predict each video is also much longer than the proposal based method in~\cite{fast-proposal}: $> 108$s.  

\end{itemize}

\subsection{ActivityNet}

There are 200 activity classes and 4926 videos in the validation set of ActivityNet. Each video has 3254 frames on average. 

\begin{itemize}
\item \textbf{Ours} is independent of video length, but is determined by the number of steps to run our policy. We take the same processing as for THUMOS14 to perform activity detection on this dataset. So the overall computation time to predict each video is the same: $(0.1\times15\times6) + (3.0\times15\times6) + (5.4\times6) + (0.1\times10) + (3.0\times10) + 5.5 \approx$ 348ms $\approx 0.35$s. 

\item \textbf{Glimpses}~\cite{yeung2015end} downsamples videos by 25 (i.e., to 1fps) for this dataset, and observes $2\%$ (or less) of video frames for inference. Similarly, we need to train and run 200 models to detect the whole classes of this dataset. So the average computation time to detect the entire classes for each video is: $3254 \div 25 \times 0.02 \times (3.0 + 8.2) \times 200 \approx 5824$ms $\approx 5.8$s.

\item \textbf{R-C3D}~\cite{r-c3d} runs at the same speed: 569fps on an NVIDIA Titan X Maxwell GPU and 1030fps on an NVIDIA Titan X Pascal GPU. So the average computation time to predict each video is: $3254 \div 569 \approx 5.7$s on Titan X Maxwell, and $3254 \div 1030 \approx 3.2$s on Titan X Pascal.

\item \textbf{OBU16}~\cite{singh2016untrimmed} uses C3D for feature extraction running at 313.9fps on the GPU, a binary random forest, and a dynamic programming process for generating proposals. So only considering the feature extraction, the average computation time to process each video is: $3254 \div 313.9 \approx 10$s. 

\item \textbf{UPC16}~\cite{upc} consists of C3D running at 313.9fps on the GPU, and a single layer LSTM predicting with 2.7ms on the GPU. C3D uses a short video clip of 16 frames as input. So the average computation time to predict each video is: $(3254 \div 313.9)$s $+$ $(3254 \div 16 \times 2.7)$ms $\approx 11$s 

\item \textbf{UTS16}~\cite{uts} extracts a variety of features such as iDT features, C3D features, deep features by ResNet152 (pre-trained on ImageNet), deep features by ResNet152 (pre-trained on Places2), and deep features by InceptionV3. Additional costs involve VLAD encoding of multiple features and SVM classification. So only considering the feature extraction by iDT which runs at 3.5fps on the CPU, the average computation time to process each video is: $3254 \div 3.5 \approx 930$s. 

\item \textbf{CDC}~\cite{DBLP:journals/corr/ShouCZMC17} relies on the detection outputs of UTS16~\cite{uts}, i.e., it is used to refine the predicted results of UTS16. While CDC runs at 500fps in the refining process, the computation bottleneck comes from generating the temporal segments by UTS16. So the overall computation time is determined by UTS16: $930$s.  

\end{itemize}

\section{Illustration of Policy Execution}
\label{sec:policy}

\begin{figure*}[h]
\centering
\includegraphics[width=\linewidth]{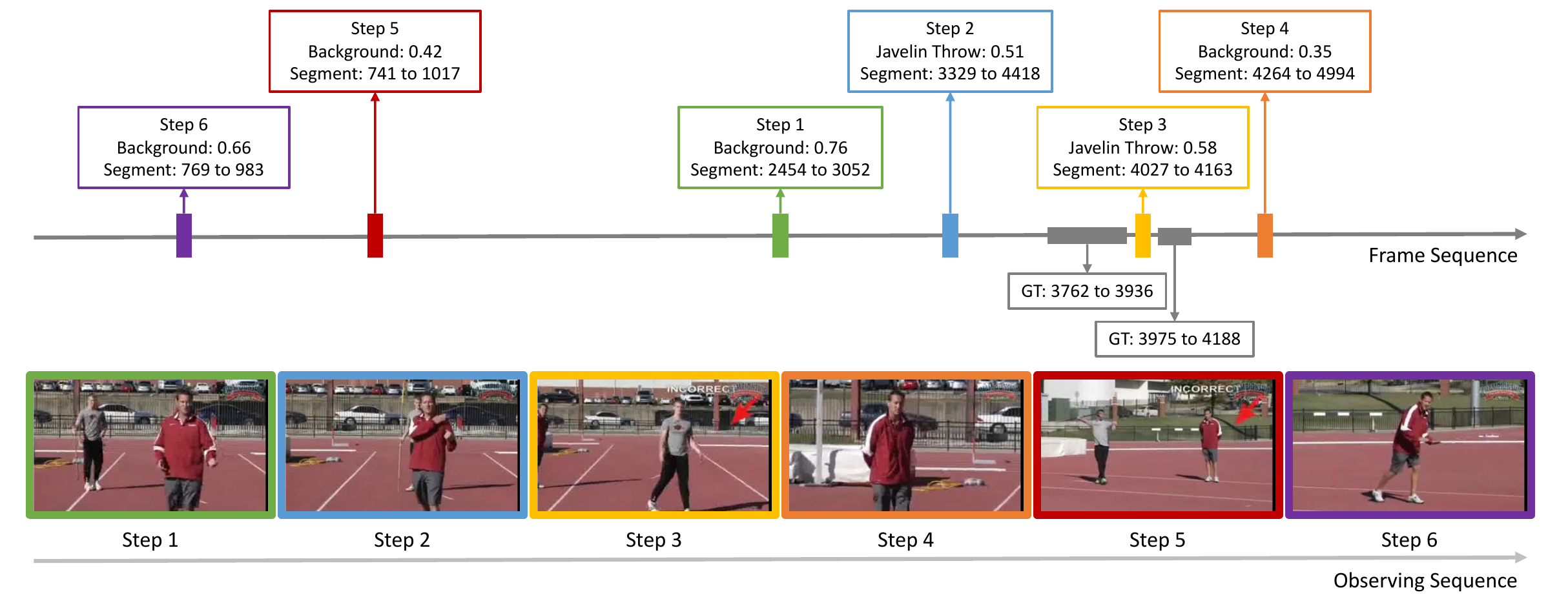}
\caption {Illustration of the learned policy running for frame selection and activity prediction on THUMOS14.}
\label{fig:policy}
\end{figure*}

Figure~\ref{fig:policy} illustrates the frame selection and prediction process of the learned policy. Each colored box above the frame sequence shows the predicted action class (with associated probability score), and detected temporal segment (from start to end). We can directly discard the segments that are predicted as background. At steps 2 and 3, the policy makes the true positive predictions that match to the two ground truth segments. Frames of the observing sequence present the selected frames for the corresponding steps.      

\section{Examples of Detected Activities}
\label{sec:examples}

\begin{figure*}[h]
\centering
\includegraphics[width=\linewidth]{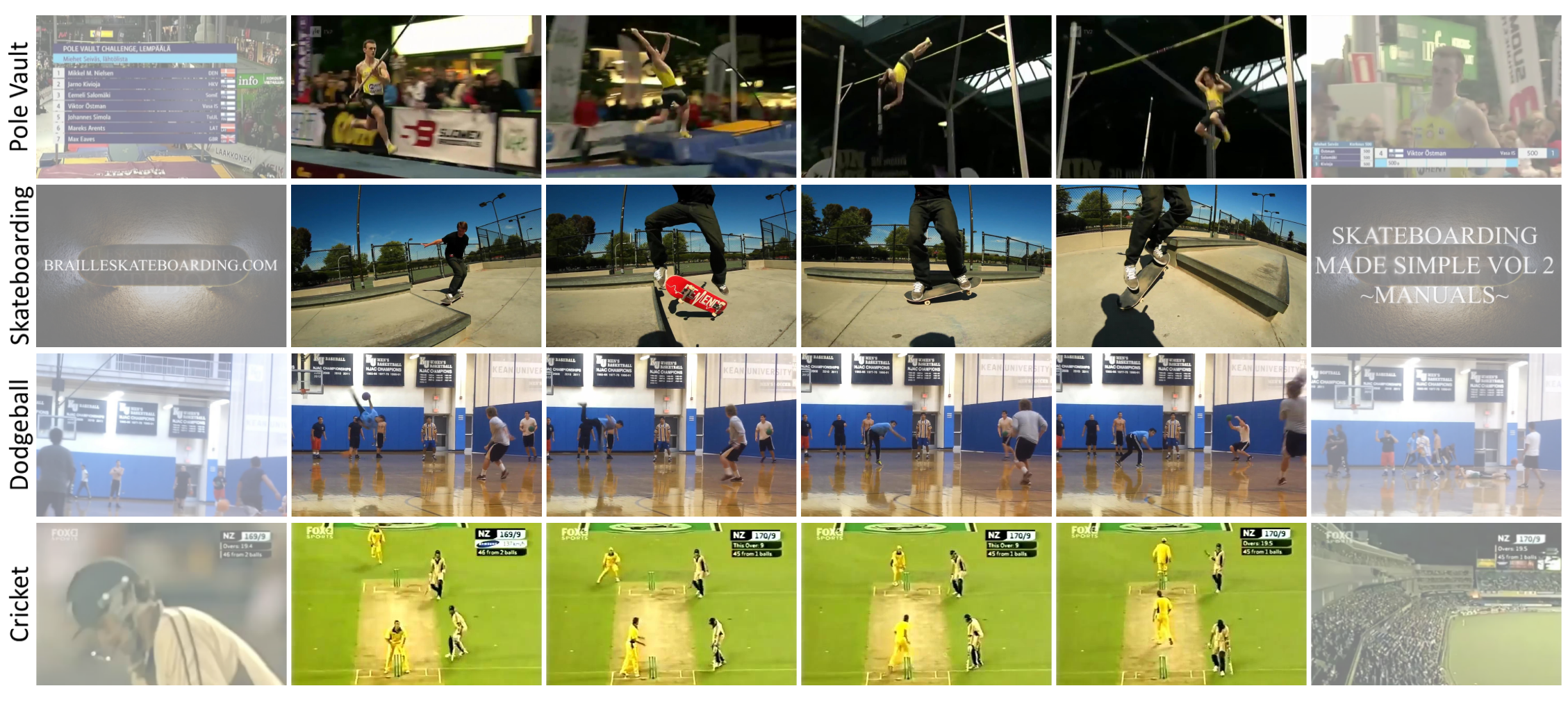}
\caption {Examples of the predicted results on ActivityNet. Each row shows four sampled frames within the temporal extent of a detected activity. Faded frames indicate the frames outside the detected temporal boundary.}
\label{fig:detected-examples}
\end{figure*}

Figure~\ref{fig:detected-examples} demonstrates the prediction examples of our model, including the challenging classes that involve great scale change, large viewpoint variations, and crowded background.     

\section{More Results on ActivityNet}
\label{sec:results}

Since the glimpses method~\cite{yeung2015end} is a binary model and their detection results on the entire 200 classes of ActivityNet are not provided (otherwise they have to train 200 models for each class), we train our policy on the same two subsets (i.e., sports and work) as~\cite{yeung2015end} for fair comparisons. As shown in Tables~\ref{tab:sports} and~\ref{tab:work}, our approach perform better than~\cite{yeung2015end} in 15 out of 21 classes on the sports subset, and 9 out of 15 classes on the work subset. Overall, we outperform~\cite{yeung2015end} by 2.0\% and 1.8\% in mAP on the sprots and the work subsets, respectively. More importantly, we need only a single training phase to handle the entire activity classes of each subset, while~\cite{yeung2015end} requires to train multiple models for every class.

\begin{table}[thb]
\small
	\centering
	\begin{tabular}{lcclcc}
		\toprule
		Class & Glimpses~\cite{yeung2015end} & Ours & Class & Glimpses~\cite{yeung2015end} & Ours\\
		\midrule
		Archery & 5.2 & \textbf{13.7} & Long Jump & \textbf{56.8} & 48.7\\
		Bowling & 52.2 & \textbf{52.4} & Mountain Climb. & \textbf{53.0} & 52.4\\
        Bungee & \textbf{48.9} & 46.3 & Paintball & 12.5 & \textbf{24.9}\\
        Cricket & 38.4 & \textbf{39.1} & Playing Kickball & 60.8 & \textbf{61.2}\\
        Curling & 30.1 & \textbf{32.3} & Playing Volleyball & \textbf{40.2} & 39.2\\
        Discus Throw & 17.6 & \textbf{21.8} & Pole Vault & 35.5 & \textbf{40.2}\\
        Dodgeball & \textbf{61.3} & 60.2 & Shot Put & 50.9 & \textbf{51.4}\\
        Doing Moto. & 46.2 & \textbf{47.3} & Skateboard. & \textbf{34.4} & 32.7\\
		Ham. Throw & 13.7 & \textbf{18.8} & Start Fire & 38.4 & \textbf{40.1}\\
        High Jump & 21.9 & \textbf{27.4} & Triple Jump & 16.1 & \textbf{22.7}\\
        Javelin Throw & 35.7 & \textbf{40.1} & & & \\
        \midrule
        mAP & & & & 36.7 & \textbf{38.7}\\
		\bottomrule 
	\end{tabular}
	\vspace{10pt}
	\caption{Comparison of the per-class breakdown AP at IoU of $\alpha = 0.5$ on the sports subset of ActivityNet.}
	\label{tab:sports}
\end{table}

\begin{table}[thb]
\small
	\centering
	\begin{tabular}{lcclcc}
		\toprule
		Class & Glimpses~\cite{yeung2015end} & Ours & Class & Glimpses~\cite{yeung2015end} & Ours\\
		\midrule
		Attend Conf. & \textbf{56.5} & 53.8 & Phoning & \textbf{52.1} & 46.7\\
		Search Security & 33.9 & \textbf{36.1} & Pumping Gas & 34.0 & \textbf{49.3}\\
        Buy Fast Food & 45.8 & \textbf{48.2} & Setup Comp. & 30.3 & \textbf{35.1}\\
        Clean Laptop & 35.8 & \textbf{38.3} & Sharp. Knife & 35.2 & \textbf{38.3}\\
        Making Copies & \textbf{41.7} & 39.5 & Sort Books & 16.7 & \textbf{33.7}\\
        Organizing Boxes & 19.1 & \textbf{24.5} & Using Comp. & \textbf{50.2} & 47.3\\
        Organizing Cabin. & 43.7 & \textbf{46.2} & Using ATM & \textbf{64.9} & 50.6\\
        Packing & \textbf{39.1} & 38.3 & & & \\
        \midrule
        mAP & & & & 39.9 & \textbf{41.7}\\
		\bottomrule 
	\end{tabular}
	\vspace{10pt}
	\caption{Comparison of the per-class breakdown AP at IoU of $\alpha = 0.5$ on the work subset of ActivityNet.}
	\label{tab:work}
\end{table}

\end{document}